\documentclass[11pt]{article}

\usepackage[preprint]{acl}

\usepackage{times}
\usepackage{latexsym}
\usepackage[T1]{fontenc}
\usepackage[utf8]{inputenc}
\usepackage{microtype}
\usepackage{tabularx}
\usepackage[table]{xcolor}  
\usepackage{graphicx}
\usepackage{booktabs}
\usepackage{multirow}
\usepackage{colortbl}
\usepackage{amsmath,amssymb,amsfonts}
\usepackage{xcolor}
\usepackage{textcomp}
\usepackage{subcaption}     
\usepackage{enumitem}
\usepackage{xurl}
\usepackage{xspace}

\DeclareUnicodeCharacter{2260}{\ensuremath{\neq}}
\DeclareUnicodeCharacter{2264}{\ensuremath{\leq}}
\DeclareUnicodeCharacter{2265}{\ensuremath{\geq}}

\setlist{leftmargin=1.0em, itemsep=2pt, parsep=0pt,
         topsep=4pt, partopsep=0pt}
\setlist[enumerate]{leftmargin=*, widest*=13, labelsep=0.35em, itemsep=2pt,
                   parsep=0pt, topsep=4pt, partopsep=0pt}

\usepackage{listings}

\lstnewenvironment{verbatim}
  {\lstset{basicstyle=\ttfamily\scriptsize,
           breaklines=true, breakatwhitespace=false,
           columns=fullflexible, keepspaces=true,
           xleftmargin=0.4em, aboveskip=5pt, belowskip=5pt}}
  {}

\newcommand{\fitwidth}[2]{%
  \resizebox{\ifdim\width>#1 #1\else\width\fi}{!}{#2}}

\newcommand{\dataset}{\textsc{Arafa}\xspace}

\newcommand{\gc}[1]{\textcolor{black}{#1}}
\newcommand{\corr}[1]{\textcolor{black}{#1}}

\providecommand{\backmatter}{}
\newcommand{\bmhead}[1]{\section*{#1}}

\usepackage{arabtex}
\usepackage{utf8}
\setcode{utf8}

\makeatletter
\protected\def\begin#1{%
  \UseHook{env/#1/before}%
  \@ifundefined{#1}%
    {\def\reserved@a{\@latex@error{Environment #1 undefined}\@eha}}%
    {\def\reserved@a{\def\@currenvir{#1}%
        \edef\@currenvline{\on@line}%
        \@execute@begin@hook{#1}%
        \csname #1\endcsname}}%
  \@ignorefalse
  \begingroup
  \let\end\a@l@end 
  \@endpefalse\reserved@a}
\makeatother

\newcommand{\AR}[1]{\par\begin{RLtext}\small #1\end{RLtext}}

\makeatletter
\renewenvironment{abstract}%
  {{\centering\large\textbf{\abstractname}\par}\vspace{0.5ex}%
   \list{}{\setlength{\rightmargin}{0.6cm}\setlength{\leftmargin}{0.6cm}}%
   \item[]\ignorespaces\@setsize\normalsize{12pt}\xpt\@xpt}%
  {\unskip\endlist}
\makeatother

\title{\dataset: An LLM-Generated Arabic Fact-Checking Dataset}

\author{
  Christophe Khalil\textsuperscript{*}, Shady Elbassuoni, Rida Assaf \\
  Computer Science, American University of Beirut, Beirut, Lebanon \\
  \texttt{cak29@mail.aub.edu}\textsuperscript{*} \quad
  \texttt{se58@aub.edu.lb} \quad
  \texttt{ra278@aub.edu.lb} \\
  \vspace{2pt}
  \textsuperscript{*}\,Corresponding author.
}

\begin{document}
\maketitle

\begin{abstract}
Automatic fact-checking poses a significant challenge in Arabic natural language processing due to the scarcity of datasets and resources. In this manuscript, we introduce \dataset, a new large-scale dataset for fact-checking in Modern Standard Arabic, constructed through an automated framework leveraging large language models (LLMs). The dataset was constructed through a three-step pipeline: (1) claim generation from Arabic Wikipedia pages with supporting textual evidence, (2) claim mutation to generate challenging counterfactual claims with refuting evidence, and (3) an automatic validation step to validate that the generated claims are either supported or refuted by their accompanying evidence, or if the evidence does not provide enough information to judge the validity of the claims. The resulting dataset comprises 181,976 claim-evidence pairs labeled as \textit{supported}, \textit{refuted}, or \textit{not enough information}. Human evaluation carried out on a test sample from the dataset demonstrated strong inter-annotator agreement ($\kappa = 0.89$) using Cohen’s Kappa for supported claims and ($\kappa = 0.94$) for refuted claims. Automatic validation based on a human-evaluated sample achieved 86\% accuracy for supported claims and 88\% for refuted ones. To showcase \dataset's value as a resource for automatic Arabic fact-checking, four open-source transformer-based models were fine-tuned using \dataset, with the top-performing model achieving a Macro F1-score of 77\% on the test data. In addition to \dataset being the first large-scale dataset for Arabic fact-checking, our framework presents a scalable approach for developing similar resources for other low-resource languages.

\end{abstract}

\section{Introduction}

The rapid spread of misinformation and false claims in today’s fast-evolving digital landscape poses significant societal risks. Misinformation can 
influence public opinion, sway electoral outcomes, and erode trust in businesses, public institutions, and scientific consensus \cite{vosoughi2018spread}. These widespread implications highlight the need for reliable automated fact-checking solutions, which serve as a protective measure to enable verification at scale, a task that remains infeasible for human fact-checkers alone. Automatic fact-checking has traditionally relied on deep learning models trained on high-quality datasets \cite{thorne2018automated}, which are typically annotated by human annotators, which is both time-consuming and resource-intensive \cite{thorne2018fever}.
The introduction of Large Language Models (LLMs) in recent years has reduced the dependence on human annotators for automatic fact-checking \cite{zhao2023survey}. Some approaches utilized these models to generate datasets \cite{pan2021zero}, while others directly employed them to perform fact-checking \cite{cheung2023factllama}.
However, most of these approaches  have primarily focused on high-resource languages such as English \cite{pan2021zero, cheung2023factllama}.

As a result, comparable resources and models for Arabic remain underexplored. Existing Arabic fact-checking datasets, including AraFacts \cite{ali2021arafacts} and AraStance \cite{alhindi2021arastance}, are limited in scope, size, and domain, as they primarily focus on news or posts from X, which hinders their ability to effectively train and evaluate general Arabic fact-checking models.
Recent advances in LLMs offer a way to address this resource gap through automated dataset generation, allowing researchers to evaluate how reliably they can produce human-quality training data and support scalable fact-checking in low-resource languages.
To address this gap, we introduce \dataset, a large and comprehensive Arabic fact-checking dataset. Unlike existing resources, \dataset is a \emph{large general-purpose} dataset designed to train deep learning models for fact-checking across domains and to serve as a benchmark for evaluating Arabic fact-checking systems. It contains \emph{181,976} natural language claims with textual evidence that either supports, refutes, or provides insufficient information (i.e., not enough information, NEI) to assess each claim. The dataset spans a diverse range of topics across multiple domains, including but not limited to humanities (history, politics, arts, culture, religion), sciences (technology, biology, environment), social sciences (society, economics, organizations), and biographical content (people, events, locations).

\begin{table*}[t!]
\centering
\small
\caption{Dataset Overview and Examples}
\begin{tabular}{@{}p{0.085\textwidth}p{0.075\textwidth}>{\raggedright\arraybackslash}p{0.375\textwidth}>{\raggedright\arraybackslash}p{0.375\textwidth}@{}}
\toprule
\textbf{Label} & \textbf{Count} & \textbf{Example Claim} & \textbf{Example Evidence}\\ \midrule

 Supported & 111,303 & \AR{أطلق نادي علمي في بيروت برنامج الفضاء اللبناني عام 1960}  & \AR{عام 1960، بدأ نادي علمي من إحدى جامعات بيروت برنامجًا فضائيًا سُمّي حينها برنامج الفضاء}  \\
 
 & & \textit{A scientific club in Beirut launched the Lebanese space program in 1960} & \textit{In 1960, a scientific club from one of Beirut's universities started a space program called 'Space Program'}  \\
\midrule
Refuted & 42,109 & \AR{يتم استخراج الألماس غالبًا من الفوهات البركانية التي تجلبه من أعماق تصل إلى 200 كيلومترًا تحت الأرض} & \AR{يستخرج معظم الألماس من الفوهات البركانية حيث تلقي به الحمم البركانية التي تحضره من أعماق الأرض من مسافات قد تصل إلى 150 كيلومترًا } \\

& & \textit{Diamonds are often extracted from volcanic craters that bring them from depths up to 200 kilometers beneath the Earth's surface} & \textit{Most diamonds are extracted from volcanic craters where lava carries them from depths up to 150 kilometers} \\ 
\midrule
NEI & 28,564 & \AR{إس كيو إل ألشمي يُستخدم بشكل شائع لتحديد البيانات العلائقية في البرمجيات} & \AR{يمكن استخدام إس كيو إل ألشمي كمحدد للبيانات العلائقية} \\
& &  \textit{SQLAlchemy is commonly used for defining relational data in software} & \textit{SQLAlchemy can be used as a relational data mapper} \\
\bottomrule
\end{tabular}
\label{tab:dataset}
\end{table*}

In line with recent approaches that leverage LLMs for automatic fact-checking \cite{pan2021zero, cheung2023factllama}, \dataset is generated using an automated framework, significantly reducing the need for manual annotation and enabling large-scale dataset creation. Our design decision is supported by the demonstrated strengths of these models in natural language understanding, contextual reasoning, and linguistic nuance detection \cite{zhao2023survey, brown2020language}. Prior studies further show that LLMs can generate diverse and accurate claims from reliable sources such as Wikipedia \cite{pan2021zero}, which is essential for producing high-quality datasets that can capture the complexity of natural language \cite{zhou2019gear, liu2019fine}. Building on these findings, \dataset adopts a fully automated construction pipeline composed of three interdependent tasks: claim generation, claim refutation, and validation.
In the generation task, we leverage GPT-4o to automate the generation of supported claims along with their evidence from Arabic Wikipedia, thereby eliminating the dependence on manual annotation. In the refutation task, a subset of these supported claims is mutated by GPT-4o to generate counterfactual versions of the claims using carefully designed techniques. Finally, in the validation task, both supported and refuted claims and their evidence are passed to Claude Sonnet 3.5 to verify the relationship between the claim and its evidence and label each pair as supported, refuted, or not enough information (NEI). Table \ref{tab:dataset} summarizes the dataset statistics and provides examples of claim-evidence pairs for each label. 

To evaluate \dataset, we followed a standard human-in-the-loop evaluation procedure used for many existing fact-checking datasets. A random sample of 200 supported claims and 200 refuted claims was annotated by two \emph{expert} annotators. The annotators \emph{collectively} labeled each claim as supported, refuted, or NEI based on its evidence. This provided ground-truth claim-evidence pairs to assess the validation LLM, which achieved an accuracy of 86\% on supported claims and 88\% on refuted claims. To compare this to human-level performance, the same ground-truth claim-evidence pairs were also annotated by four other human annotators. The non-expert annotators achieved strong inter-annotator agreement (Cohen's $\kappa = 0.89$ for supported claims and $\kappa = 0.94$ for refuted claims) and high agreement with the expert annotations ($\kappa = 0.86$ and $0.84$ respectively). This highlights that LLM-validated labels are sufficiently reliable and consistent with human judgment.

To the best of our knowledge, \dataset is the largest Arabic fact-checking dataset, which is publicly available\footnote{\url{https://github.com/chriskhalil/ARAFA}}. We believe that this dataset will serve as a valuable resource to advance automatic fact-checking for the Arabic language by serving as both training data and a standard benchmark. To demonstrate this, we fine-tuned four open-source transformer-based models using \dataset, with the best-performing model AraModernBert-Base-V1.0 achieving a Macro F1-score of 77\% on test data, confirming that \dataset is indeed suitable for training a robust Arabic fact-checking approach. Moreover, our framework for generating \dataset using LLMs presents a scalable approach for developing similar fact-checking datasets for other low-resource languages. 

\section{Related Work} 
\label{sec:lit}
Automatic fact-checking has evolved in recent years, transforming how we verify information in the digital age. Early systems used multi-stage pipelines that combined information retrieval (IR) modules with verification components \cite{hanselowski2019richly, nie2019combining}. These systems typically retrieve evidence from external knowledge sources such as Wikipedia and then use that evidence to assess the veracity of claims. These early studies highlighted the critical role of both effective evidence retrieval and rigorous claim verification using the retrieved evidence for robust fact-checking \cite{hanselowski2019richly, nie2019combining}. A significant milestone in the field was the introduction of the FEVER dataset \cite{thorne2018automated}. FEVER, notable for its scale and breadth of topics it covers, provided claims paired with evidence sentences, categorized as \textit{supported}, \textit{refuted}, or \textit{not enough information}, all derived from Wikipedia. The FEVER dataset has influenced numerous subsequent studies and research in the area of fact-checking. For instance, e-FEVER \cite{stammbach2020fever} built upon FEVER by adding abstract summaries of the  evidence, thus providing human-readable explanations of how the evidence supports or refutes claims. Another extension, FEVEROUS \cite{aly2021feverous}, broadened the scope to include unstructured text and structured information like tables, presenting more complex and realistic fact-checking scenarios. More recently, EX-FEVER \cite{ma2024ex} introduced a large-scale dataset of over 60,000 multi-hop claims with human-annotated explanations, which introduced complex reasoning in fact-checking tasks.

The field further advanced with the advent of Large Language Models (LLMs), leading to extensive research on their integration within fact-checking workflows \cite{zhao2023survey, zhang2023towards, aly2021feverous}. Building on this, the QACG framework was introduced \cite{pan2021zero}, which is a zero-shot fact-checking approach. QACG generates claims automatically from given evidence using question-answer pairs, reducing the need for human-annotated claims and broadening the variety of claim types fact-checking models can handle. Another work \cite{lee2020language} explored the capabilities of pre-trained LLMs as end-to-end fact checkers, demonstrating their ability to leverage knowledge acquired during pre-training to verify facts, sometimes even without relying on external evidence sources. However, this approach also raised concerns about the transparency and potential for hallucination or fabrication of evidence by the LLM, as the source of information on which it judges the validity of claims is opaque. To this end, one approach \cite{liu2019fine} investigated hierarchical step-by-step prompting methods to improve the explainability and performance of LLMs in news fact-checking. More recent work \cite{wang2023explainable} looked at how LLMs can improve transparency in fact-checking by using knowledge-grounded reasoning. Finally, FactLLaMA \cite{cheung2023factllama} explored how to optimize LLMs for fact-checking through methods like instruction tuning with external knowledge, enhancing their fact-checking abilities by grounding them in knowledge bases.

In more recent developments, some datasets leveraged the power of generative AI for synthetic data generation; for instance, MultiSynFact \cite{chung2025beyond} presented a large-scale synthetic multi-lingual dataset containing over 2.2 million claim-evidence pairs across Spanish, German, and English. Their work demonstrated that automated synthetic data for scarce resources can significantly improve fact-checking models' performance, validating the feasibility of such pipelines.
\begin{table*}[t!]
\centering
\small
\caption{Comparison of \dataset with Major Fact-Checking Datasets, N/R: Not Reported}
\label{tab:dataset-comparison}
\begin{tabularx}{\textwidth}{>{\raggedright\arraybackslash}X
                             >{\centering\arraybackslash}X
                             >{\centering\arraybackslash}X
                             >{\centering\arraybackslash}X}
\toprule
\textbf{Dataset} & \textbf{Language} & \textbf{Size} & \textbf{Evidence Source} \\
\midrule
\multicolumn{4}{c}{\cellcolor{blue!5}\textit{\textbf{Manually Annotated Datasets}}} \\
\midrule
FEVER & English & 185,445 & Wikipedia \\
e-FEVER & English & 185,445 & Wikipedia + summaries \\
FEVEROUS & English & 87,026 & Wikipedia (text+tables) \\
EX-FEVER & English & 60,000+ & Wikipedia \\
CHEF & Chinese & 10,000 & Fact-checking websites \\
CFEVER & Chinese & 30,012 & Wikipedia \\
SciFact & English & $\sim$1,400+ & Scientific abstracts \\
PubHealth & English & 11,832 & Fact-checking websites \\
AraFacts & Arabic & $\sim$6,222 & News articles \\
AraStance & Arabic & $\sim$4,063 & Twitter/X posts \\
AuRED & Arabic & $\sim$6,000 & Twitter + sources \\
AraFactEx & Arabic & $\sim$500 & News/rumors \\
\midrule
\multicolumn{4}{c}{\cellcolor{blue!5}\textit{\textbf{LLM-Generated Datasets}}} \\
\midrule
QACG & English & 795,746 & Wikipedia \\
FactLLaMA & English & $\sim$10,000+ & Knowledge bases \\
MultiSynFact & Multi (ES/DE/EN) & 2,200,000 & Wikipedia \\
\midrule
\textbf{\dataset (Ours)} & \textbf{Arabic (MSA)} & \textbf{181,976} & \textbf{Wikipedia} \\
\bottomrule
\end{tabularx}
\end{table*}
In addition to general-domain fact-checking, domain-specific datasets were developed. SciFact \cite{wadden2022scifact} was specifically developed for claims that require expert knowledge in the biomedical domain. It introduced 1,409 scientific claims verified against 5,183 abstracts from biomedical literature.
PubHealth \cite{kotonya2020explainable}, another domain-specific dataset, contains 11,800 claims with explanations that target the public health sector, where misinformation can have serious health consequences. Efforts to develop fact-checking resources in languages other than English have produced datasets such as CHEF \cite{hu2022chef}, which is one of the first Chinese multi-domain fact-checking datasets, comprising 10,000 real-world claims with manually annotated evidence retrieved from the internet. Building on this, CFEVER \cite{lin2024cfever} presented a larger Chinese dataset of 30,012 manually annotated claims based on Chinese Wikipedia, following the FEVER methodology.

While English fact-checking has seen substantial advancements, empowered by large-scale datasets and the adaptation of LLMs, Arabic fact-checking remains in its early stages. Existing Arabic fact-checking datasets such as AraFacts \cite{ali2021arafacts} and AraStance \cite{alhindi2021arastance}, while valuable, are considerably small and often limited to specific domains like news or social media, representing only a fraction of the data volume available in Arabic. Other Arabic fact-checking datasets include AuRED \cite{haouari2024aured}, which focuses on Arabic rumor verification using evidence from Twitter and other sources, and AraFactEx \cite{althabiti2024ta}, a news and rumor based dataset of limited size. The scale and scope of these Arabic fact-checking datasets highlight the crucial need for larger, more diverse datasets that capture the linguistic and cultural nuances of the Arabic language, and can be used to train or evaluate robust automatic fact-checking approaches.

Our dataset aims to bridge this gap by being the first \emph{publicly available} large and comprehensive (Modern Standard) Arabic fact-checking dataset. Table \ref{tab:dataset-comparison} provides a comprehensive comparison of \dataset with major fact-checking datasets across different languages and construction methodologies. We addressed the limitations of existing Arabic datasets in terms of scale and scope by utilizing LLMs to eliminate the need for manual human annotations and to capture a diverse set of topics. While our dataset serves as a valuable resource for training and evaluating Arabic fact-checking models, our framework for the generation of the dataset can also be used to generate similar datasets for other low-resource languages with relatively simple adaptations. 

\section{Dataset Generation}
\label{sec:app}

\begin{figure}[t]
    \centering
    \includegraphics[width=0.48\textwidth]{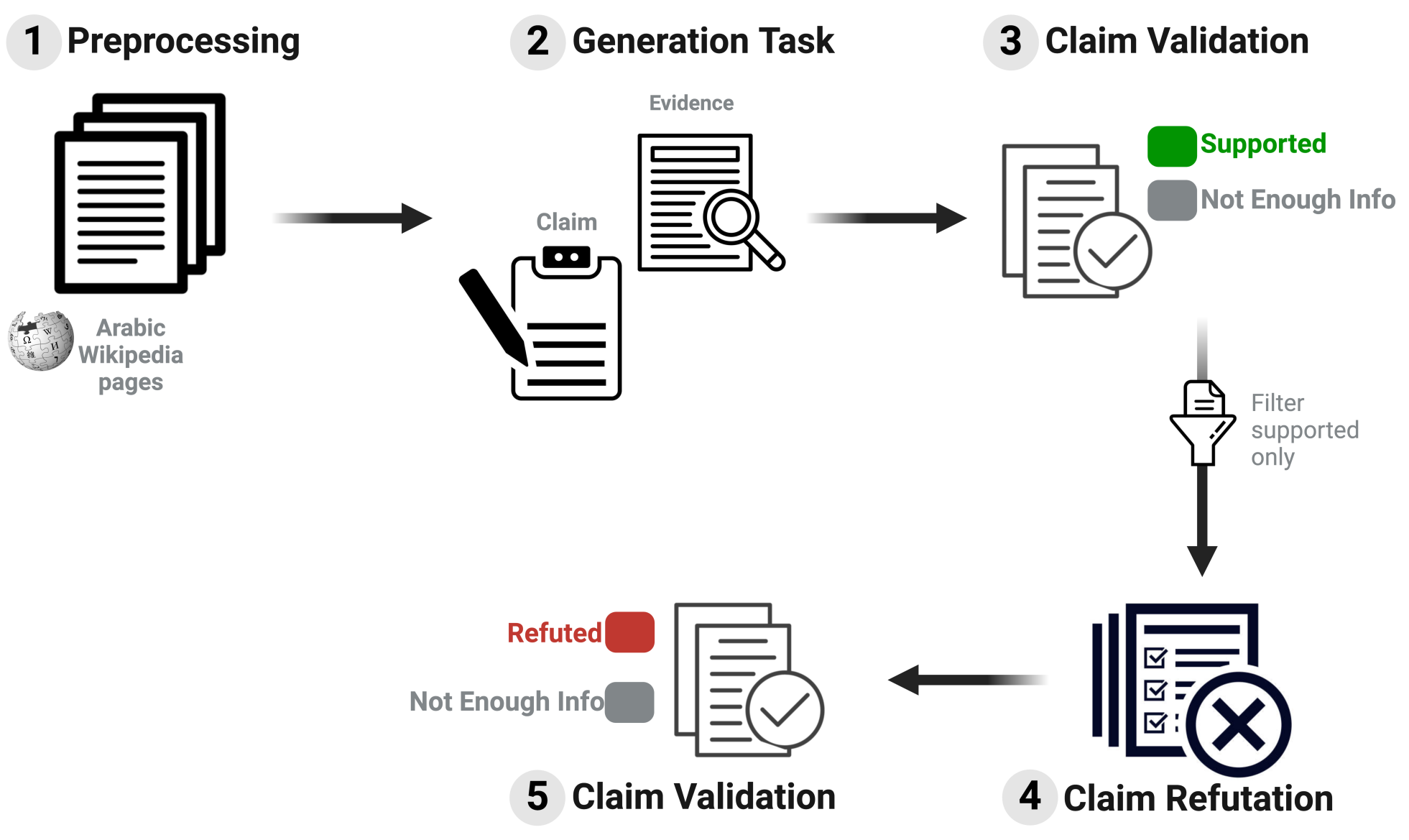}
    \caption{Dataset Generation Framework}
    \label{fig:method}
\end{figure}

\dataset was generated using a rigorous multitask framework, which is outlined in Figure \ref{fig:method}. The dataset generation process started by pre-processing the Arabic Wikipedia dump\footnote{\href{https://dumps.wikimedia.org/arwiki/}{Arabic Wikipedia Dump}} to remove irrelevant content, such as category tags, internal and external links, templates, and other artifacts. Around 10,000 clean pages were then randomly sampled, covering diverse topics, including humanities (history, politics, arts, culture, religion), sciences (technology, biology, environment), social sciences (society, economics, organizations), and biographical content (people, events, locations). The sampled pages were then chunked using a paragraph-aware chunking strategy that maintains contextual coherence, thereby mitigating the information loss typically associated with conventional text-splitting techniques that overlook natural paragraph boundaries. 

Using a carefully designed prompt, a generative model processed each chunk to generate sets of claims paired with their supporting textual evidence. To ensure the claims are supported, we employed the validation LLM to verify the relation between each claim and its evidence and label each pair as either supported or NEI when evidence was insufficient. A subset of the supported claims was then passed to the refutation task, which applied sophisticated subtle mutations to generate counterfactual claims that are refuted by the evidence. These mutated claims were again verified by the validation LLM to confirm that they were indeed refuted or marked as NEI if the evidence was inconclusive. We describe each task in more detail in the following sections. 

\subsection{Claim Generation Task}
\begin{figure}[t]
    \centering
    \includegraphics[width=0.48\textwidth]{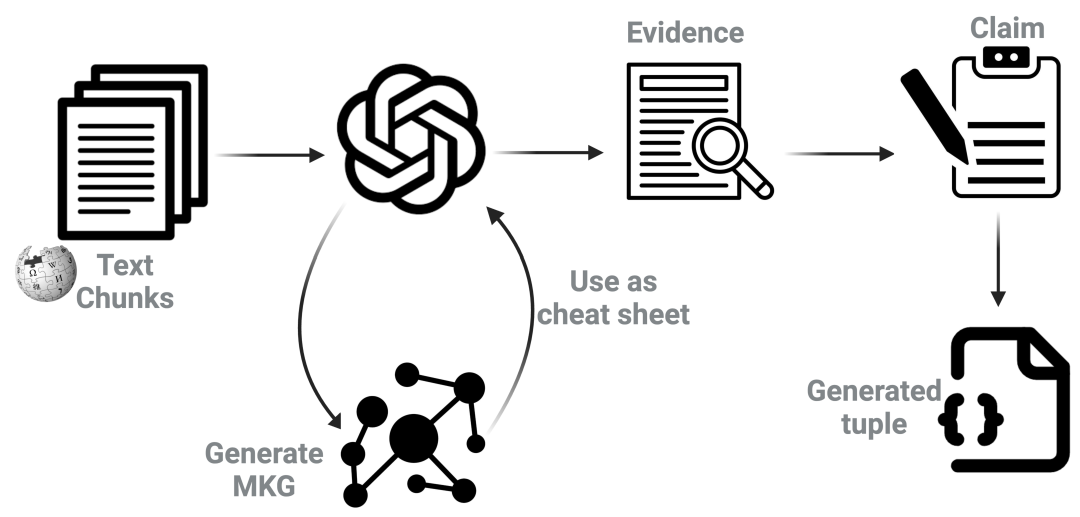}
    \caption{Claim Generation Task}
    \label{fig:generation}
\end{figure}

The goal of this task is to produce high-quality claim-evidence pairs from Arabic Wikipedia articles without manual annotation, using an LLM (Figure \ref{fig:generation}). Each text chunk was processed by the model which identified text spans as evidence and generated claims supported by each extracted span. An initial zero-shot prompt approach, which was based on the human annotation guidelines \cite{thorne2018fever}, revealed significant limitations. First, the LLM showed a tendency to extract duplicate or near-duplicate claim-evidence pairs from a given chunk. Second, due to the complexity of the task, the model showed a high rate of hallucinations in both claim generation and evidence extraction. Instead of generating claims based on the evidence, it often paraphrased the extracted evidence to align with the claims. In other cases, it relied on its pre-training knowledge to augment the generated claims with auxiliary information that was not present in the evidence. Third, coreference resolution was unreliable. Both the claim and the evidence were referencing ambiguous entities and both must be self-contained. Fourth, the grammar and the linguistic style of the generated claims were often poor. Even though LLMs are typically trained on vast, multilingual corpora, their performance can be noticeably uneven across languages \cite{bender2021dangers}. In particular, these models excel in English while struggling with other languages, a disparity primarily attributed to the imbalanced nature of the training data \cite{arnett2024language, jung2024understanding}. For example, OpenAI’s GPT-3 was developed using a dataset composed of 93\% English data and only 7\% for all other languages\footnote{\href{https://github.com/openai/gpt-3/blob/master/dataset_statistics/languages_by_word_count.csv}{GPT-3 dataset statistics}}, underscoring the heavy bias toward English. 

\begin{table*}[t]
    \centering
        \caption{An Example Supported Claim with Coreference Resolution}
    \small
    \begin{tabular}{@{}p{0.95\textwidth}@{}}
        \toprule
        \textbf{\textit{Original Chunk}} \\
        \AR{القرنة السوداء هي أعلى قمة في بلاد الشام وتقع في جبل المكمل في شمال لبنان. يبلغ ارتفاع القمة عن سطح البحر 3093 م.}\\
        \textit{Qarnat as-Sawda is the highest peak in the Levant and is located in Jabal al-Makmal in northern Lebanon. The peak is 3093 m above sea level.} \\
        \midrule
        \textbf{\textit{Claim}} \\
        \AR{ترتفع القرنة السوداء 3093 مترًا فوق سطح البحر} \\
        \textit{Qarnat as-Sawda rises 3093 meters above sea level} \\
        \midrule
        \textbf{\textit{Evidence}} \\
        \AR{يبلغ ارتفاع القمة عن سطح البحر 3093 م} \\
        \textit{The peak's elevation above sea level is 3093 m} \\
        \midrule
        \textbf{\textit{Entity Coreferences}} \\
        \begin{tabular}{@{}l@{\hspace{1.5em}}r@{\hspace{1em}}l@{}}
        Entity in Claim: & \RL{القرنة السوداء} & (\textit{Qarnat as-Sawda}) \\
        Coreference in Evidence: & \RL{القمة} & (\textit{the peak}) \\
        Coreference in Chunk: & \RL{القرنة السوداء} & (\textit{Qarnat as-Sawda}) \\
        \end{tabular} \\
        \bottomrule
    \end{tabular}
    \label{tab:coreference}
\end{table*}

An additional challenge when particularly dealing with Modern Standard Arabic is its intrinsic linguistic complexity. Unlike English, which has a relatively straightforward morphology and a vocabulary size ranging from 170,000 to 600,000 words\footnote{\href{https://en.wikipedia.org/wiki/List_of_dictionaries_by_number_of_words}{Languages dictionaries by number of words}}, Arabic features a rich morphological system yielding an estimated 12.3 million words. It follows strict grammatical norms, particularly regarding subject-verb agreement and adjective–noun agreement in terms of gender and number. Moreover, the language’s intricate system of subordinate clauses (including relative, conditional, temporal, and causal constructs) and its diverse use of coordinating conjunctions further amplify its grammatical complexity. This lexical richness not only highlights the distinctive nature of Arabic, but also necessitates specialized processing techniques to handle its complexity in generative tasks. These linguistic features pose significant challenges for language models. Recent research \cite{goldberg2019assessing} shows that while models like BERT can learn subject–verb agreement in English, their performance largely depends on the frequency of word patterns observed during training rather than grammatical understanding. This limitation is even more pronounced in Arabic, where complex morphology and a flexible word order hinder the model’s ability to generalize grammatical rules. Consequently, when certain syntactic structures are underrepresented in the training data, LLMs may produce inconsistent agreement patterns. This was evident in the claims generated by the zero-shot prompt, which often had poor linguistic style and lacked grammatical agreement between nouns, adjectives and verbs when it comes to gender, number or tense. 

To address all the aforementioned issues, we adopted a few-shot chain-of-thought (COT) prompt, guiding the model to break down the complex claim generation task into a series of intermediate steps, closely mimicking human-like behavior and reasoning. The COT prompt first instructs the LLM to skim through the provided text chunk to get an overall understanding of the topic, followed by an attentive reading to identify the main named entities. Next, it establishes connections between the identified entities, their relationships, coreferences, and any hierarchical structures. These connections form what we refer to as the Mental Knowledge Graph (MKG), which is a mini knowledge graph generated within the model's context window that serves as an internal "cheat sheet" to guide the claim generation process. Using the MKG, the LLM then extracts multiple \emph{verbatim} evidence texts from the given chunk. To maintain fidelity, the model extracts each evidence span exactly as it appears even if it includes spelling errors or grammatical mistakes, without modification. Once evidence was extracted, the LLM generates a complex claim that is fully supported by its corresponding evidence and extends beyond a surface-level rephrasing. It also ensures that the claim is semantically sound and grammatically correct by referring to a set of Arabic syntactic and semantic rules provided to it in the prompt (Arabic Rules\footnote{\href{https://kalimah-center.com/advanced-arabic/}{Arabic Rules}}). To ensure coherence, claims were required to focus on a single aspect of one target entity explicitly mentioned in the evidence, and to reference the target entity directly as it appears. For coreference resolution, we leveraged the MKG and the original chunk from which the evidence was extracted to map mentions to their named entities in the claim or the evidence. Finally, the LLM performed a series of validation checks, listed below, using the MKG and the original chunk to ensure quality:
\begin{enumerate}
\item Verify each claim meets all the criteria for advanced level in Arabic.
\item Ensure no internal knowledge, inference or assumptions were used.
\item Cross-check each claim against its evidence to ensure full support and accuracy.
\item Revise or extract new evidence if the claim cannot be fully supported by the evidence.
\item Make certain each claim is not a verbatim copy or simple paraphrasing of its evidence; revise to make it more complex and distinct.
\item Validate all extracted evidence spans are unique (cosine similarity = 0).
\item Inspect all generated claims per chunk are unique (cosine similarity below 0.80).
\end{enumerate}

The full COT prompt is provided in the supplementary material. This task produces a set of tuples, each containing a claim, its supporting evidence, and all entity coreferences within the claim, evidence and the source text chunk. Table \ref{tab:coreference} illustrates one such tuple alongside the original chunk from which it was derived. 
\subsection{Claim Refutation Task}

\begin{table*}[t]
\centering
\caption{Example Mutated Claims Generated Using Different Mutation Strategies}
\label{tab:mutations}
\small
\begin{tabular}{@{}>{\raggedright\arraybackslash}p{0.17\textwidth}>{\raggedright\arraybackslash}p{0.385\textwidth}>{\raggedright\arraybackslash}p{0.385\textwidth}@{}}
\toprule
\textbf{Strategy} & \textbf{Original Claim} & \textbf{Mutated Claim} \\
 \midrule
Temporal Nuance & 
\AR{افتتحت جامعة بيروت العربية عام ٠٦٩١}
 & 
\AR{افتتحت جامعة بيروت العربية عام ٠٧٩١}\\
&\textit{The Beirut Arab University was established in 1960} &
\textit{The Beirut Arab University was established in 1970} \\
\midrule
Quantitative Distortion & 
\AR{يبلغ ارتفاع جبل لبنان ٨٨٠٣ متراً}
 & 
\AR{يبلغ ارتفاع جبل لبنان ٠٠٥٣ متراً}\\
&\textit{Mount Lebanon reaches a height of 3,088 meters} &
\textit{Mount Lebanon reaches a height of 3,500 meters} \\
\midrule
Qualitative Shift & 
\AR{تتميز بيروت بمناخها المعتدل}
 & 
\AR{تتميز بيروت بمناخها الحار جداً}\\
&\textit{Beirut is characterized by its moderate climate} &
\textit{Beirut is characterized by its very hot climate} \\
\midrule
Relationship Reconfiguration & 
\AR{طرابلس هي ثاني أكبر مدينة في لبنان}
 & 
\AR{طرابلس هي أكبر مدينة في لبنان}\\
&\textit{Tripoli is the second largest city in Lebanon} &
\textit{Tripoli is the largest city in Lebanon} \\ 
\midrule
Contextual Reframing & 
\AR{يقع قصر بيت الدين في جبل لبنان}
 & 
\AR{يقع قصر بيت الدين في شمال لبنان}\\
&\textit{Beiteddine Palace is located in Mount Lebanon} &
\textit{Beiteddine Palace is located in North Lebanon} \\
\bottomrule
\end{tabular}
\end{table*}

Similar to other standard fact-checking datasets, \dataset includes false claims paired with evidence that clearly refutes them. To achieve this, we sampled 50,000 supported claims generated using the claim generation task and applied the refutation pipeline to produce challenging counterfactual claims that are contradicted by the original evidence. This approach mirrors the construction of other datasets, albeit previously done manually, and follows the intuition that refuting supported claims is more challenging than generating false claims from scratch, both for humans and LLMs \cite{thorne2018fever}. In the refutation task, each supported claim tuple, including the original claim, its supporting evidence, and all entity coreferences, was an input to the refutation pipeline, which mutated the claim into a refuted one while maintaining a sophisticated linguistic style. Rather than simply negating statements, the LLM applied carefully designed mutation strategies that produce natural and challenging refuted claims \cite{huq2020adversarial,ribeiro2018semantically}, including modifying temporal aspects, quantitative and qualitative details, entity relationships, or contextual information, while preserving grammatical correctness and plausibility. This approach ensured that the refuted claims remained on topic and semantically aligned with the original evidence. The output of the refutation task is a new tuple containing the refuted claim, along with its corresponding evidence and entity coreferences. For instance, to highlight the importance of applying these mutation strategies, consider the statement \textit{``Lebanon overlooks the Mediterranean Sea from the west''}:
\AR{يطل لبنان من جهة الغرب على البحر الابيض المتوسط}
\noindent When the LLM refutes a claim using a zero-shot prompt, without guidance from the mutation strategies, it produces \textit{``Lebanon overlooks the Red Sea from the east''}:
\AR{يطل لبنان من جهة الشرق على البحر الاحمر}
\noindent However, if the evidence does not mention Lebanon’s eastern border, this claim cannot be verified as refuted by the given evidence \emph{alone} anymore, since it is unclear whether Lebanon actually overlooks the Red Sea from the east based solely on the provided text.



Building on this approach, we applied five mutation strategies to refute claims, which were inspired by previous datasets \cite{thorne2018fever}, and adversarial text attacks \cite{huq2020adversarial,ribeiro2018semantically}, but modified to meet the particular subtleties of Arabic. 
\begin{itemize}  
    \item \textbf{Temporal Nuance:} Modifies temporal aspects of a claim while maintaining the correct Arabic grammatical structure. This includes respecting Arabic’s intricate system of temporal markers and tense agreements to ensure natural linguistic flow.  
    \item \textbf{{Quantitative Distortion:}} Alters numerical values or quantities while preserving the original claim’s semantic scope. This strategy carefully handles Arabic’s complex numerical system, including proper agreement rules and plural forms.  
    \item \textbf{{Qualitative Shift:}} Transforms descriptive attributes of entities while adhering to Arabic’s strict adjective-noun agreement rules, ensuring consistency in gender, number, and definiteness.  
    \item \textbf{{Relationship Reconfiguration:}} Modifies relationships between entities by restructuring verbal or relational constructs, while preserving the grammatical integrity of Arabic verb patterns and their syntactic implications.  
    \item \textbf{{Contextual Reframing:}} Introduces broader contextual shifts so that the modified claim becomes directly refutable by the claim's evidence.  
\end{itemize}  
Table \ref{tab:mutations} provides example mutated claims generated by applying the different mutation strategies outlined above.

Similar to the claim generation task, our claim refutation task utilized a thorough COT prompt that guided the LLM to apply the above-outlined mutation strategies to produce false claims that are non-trivially refuted by their evidence. We also enforce grammatical and contextual correctness by incorporating the same linguistic rules used during claim generation. The complete COT prompt is provided in the supplementary material.

\subsection{Large Language Model Selection and Parameter Tuning}
To carry out the claim generation and refutation tasks described above, we experimented with three major models: GPT-4o, Claude Sonnet 3.5, and Llama 3.1 70B. Note that these were considered state-of-the-art LLMs at the time the dataset was generated. Each model underwent a rigorous evaluation based on three main criteria, namely generation quality, grammatical accuracy and semantic coherence. In the pre-selection process, we conducted an in-depth qualitative analysis of the output generated. The findings of this initial analysis showed a notable difference among the three models, especially in terms of grammatical correctness and structural coherence. Llama 3.1 70B struggled with grammatical correctness and precision despite its 70 billion parameters. The model sometimes generated out-of-context sentences and displayed recurring grammatical mistakes, despite all the efforts done with the grammar injection rules and prompt engineering. These limitations led to its disqualification before a human evaluation phase took place. 

After the initial screening, the remaining two models, Claude Sonnet 3.5 and GPT-4o, underwent a human evaluation. To evaluate the quality of claims and refutations generated by the two models, we conducted multiple human evaluation surveys, each specifically designed for a single task (claim generation or refutation). Each survey consisted of a series of evidence–claim pairs, where annotators were presented with the same evidence and two competing outputs (C1 and C2), each produced by a different model. The order of presentation was randomized to avoid positional bias. Eight annotators were presented with 20 evidence and two claims per evidence, where one claim was generated by GPT-4o and one by Claude Sonnet 3.5. The annotators were asked to choose which claim they believed was better when it comes to syntactic and semantic quality, and the ability to judge its validity based on the provided evidence. For each item, annotators were asked the following question:
“Given the following C1 and C2 and evidence E, which claim do you think is better?”
The available responses were: C1 is better, C2 is better, Equally good, Neither.

\begin{figure}[t]
    \centering
    \includegraphics[width=\columnwidth]{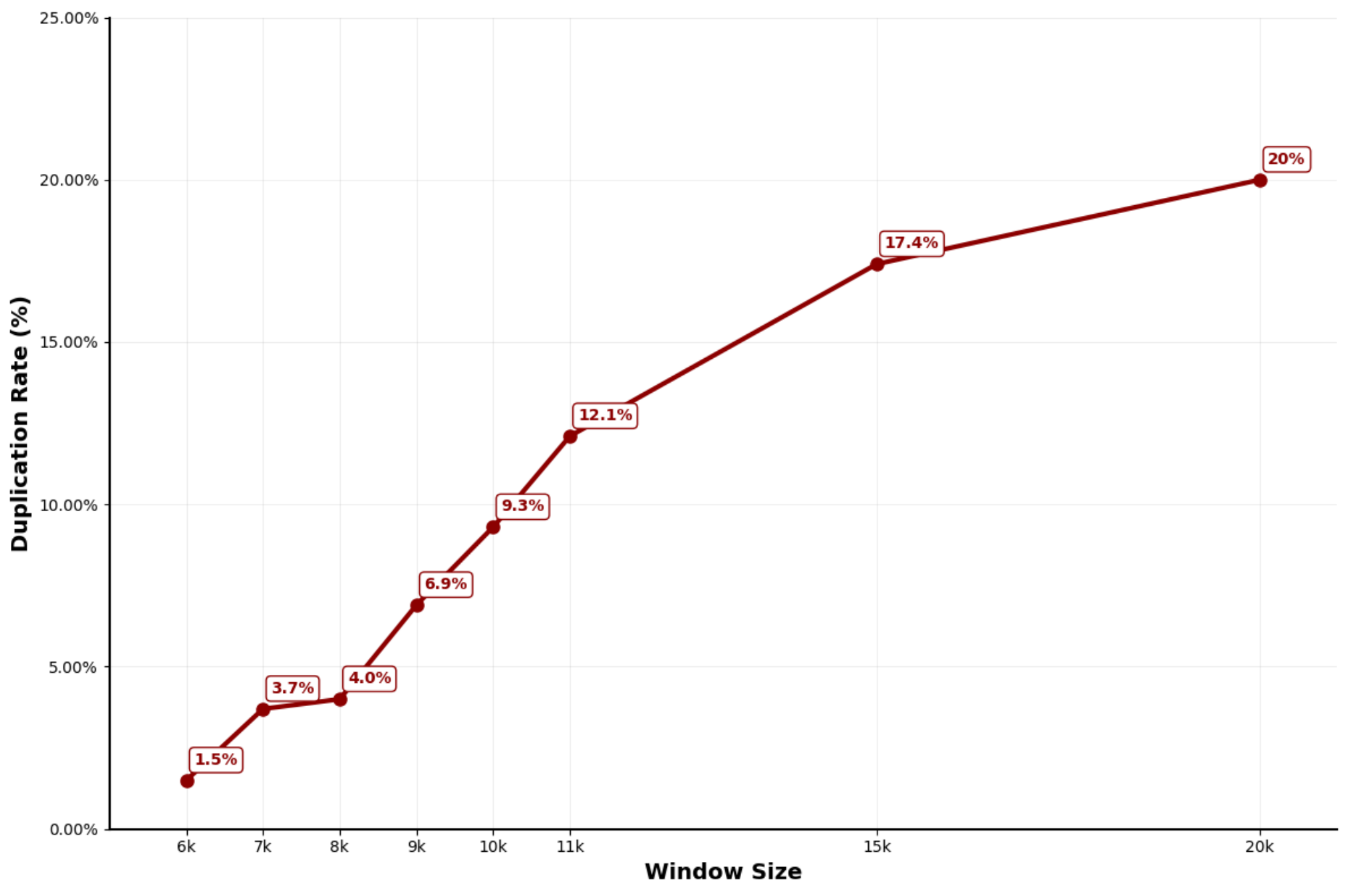}
    \caption{Rate of Duplicate Evidence Across Different Wikipedia Page Chunk Sizes}
    \label{fig:duplication-rate}
\end{figure}

We evaluated inter-annotator reliability by computing Fleiss' Kappa ($\kappa$) for both evaluation tasks. For each task, the annotators evaluated 20 evidence–claim pairs (160 total judgments each). The generation Fleiss Kappa shows slight agreement with $\kappa = 0.155$, while the refutation reveals less than chance agreement with $\kappa= -0.07$. These relatively low scores are consistent with prior research on human evaluation of natural language generation (NLG), which highlights the inherent subjectivity and low inter-annotator agreement in such tasks \cite{amidei2019use,howcroft2020twenty}. These low $\kappa$ values do not imply unreliable annotation from human annotators, but a case where both models produced outputs of comparable quality, making the distinction difficult.

Both Claude Sonnet 3.5 and GPT-4o generated high-quality claims based on human evaluation as pointed out by the annotators. GPT-4o demonstrated greater accuracy when it comes to both the generation and refutation tasks, outperforming Claude Sonnet 3.5, which tended to be more verbose or unnecessarily complex, leading to claims that were not fully supported or fully refuted by their evidence. More precisely, the human evaluation indicated that the annotators preferred the claims generated by GPT 40\% of the time compared to only 25\% for Claude, while judging them equally good in 30\% of cases and inadequate for both models in only 5\% of cases. For the refutation task, GPT was preferred 58.3\% of the time, while Claude was preferred 33.3\% of the time. Notably, the annotators never found the two models equally good (0\%), and they found neither refutation adequate only 8.4\% of the time.

Based on the human evaluation, GPT-4o was selected as the underlying LLM for both the generation and refutation tasks. The model temperature was set to 0 to reduce the randomness and ensure the verbatim extraction of evidence.
To ensure the reliability of the generation pipeline and successful completion of the given task, we conducted two empirical analyses focusing on both the input window size and the output density.
Despite GPT-4o's support for a large 120,000 context window, our experiments show that the model suffers from performance degradation as the chunk size increases due to the quadratic scaling of the attention layers \cite{du2025context,hsieh2024ruler}. As shown in Figure \ref{fig:duplication-rate}, using 20,000-token chunks resulted in a 20\% duplication rate. Reducing the chunk size to 8,000 tokens lowered this to 4\%, achieving a balance between contextual breadth for the given topic and extraction of unique verbatim evidence. 
With the window size set to 8000 tokens, we performed multiple experiments for the output density i.e. the number of evidence-claim pairs ($N$). We tested for multiple values in the range $N \in \{10,40\}$. For $N$ lower than 21, for example $N = \{10, 15\}$, we under-utilize the context window of the model, this leaves a large portion of the context unused, leading to inefficient use of the window and unnecessary API cost due to repeated transmission of partially used chunks. Conversely, for $N > 21$, we observed a decline in model performance. Specifically, in hallucination patterns that the model started to exhibit, such as combining unrelated sentences or repeating previously extracted evidence with minor paraphrasing to meet the requested count.
To balance the trade-off between both the information density and the API cost we empirically set $N=21$.

\subsection{Claim Validation Task}

Despite the rigorous prompts that we used in the claim generation and refutation tasks, it is inevitable that the LLM would not achieve 100\% accuracy on either task. This is a limitation for any automated approach that utilizes LLMs for data annotation, and in fact it is also the case when relying on human annotations \cite{thorne2018fever}. 
In the validation pipeline, we employed an LLM-as-a-judge approach to verify whether generated claims were indeed supported or refuted by their evidence. In case the validation LLM deemed the evidence to be insufficient, it output ''not enough information'', which was then used to adjust the label of the claim-evidence pair from supported or refuted to NEI. 
 
\begin{figure*}[t!]
   \centering
   \begin{subfigure}[b]{0.46\textwidth}
       \centering
       \includegraphics[width=\textwidth]{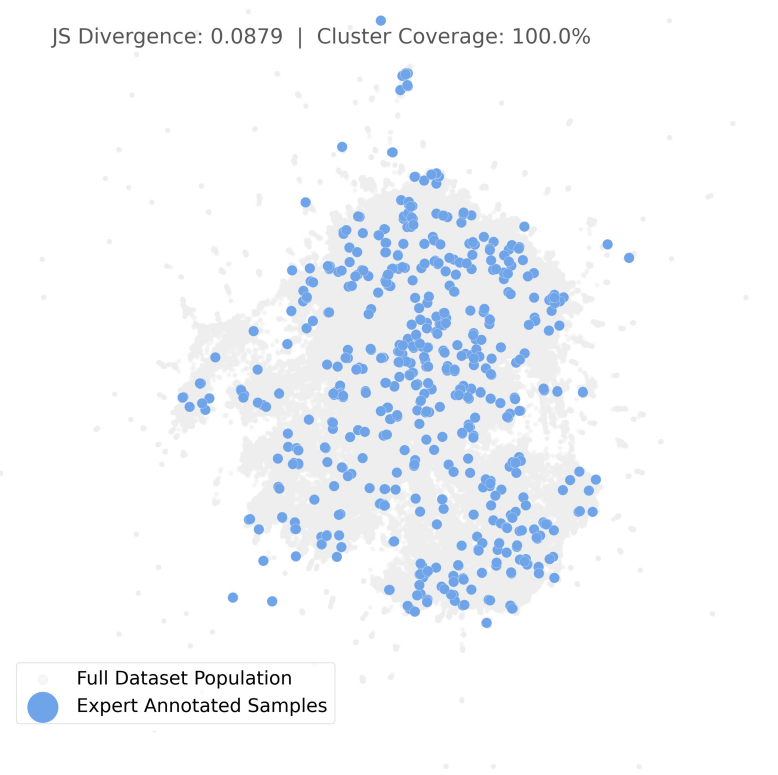}
       \caption{Semantic representativeness map}
   \end{subfigure}
   \hfill
   \begin{subfigure}[b]{0.46\textwidth}
       \centering
       \includegraphics[width=\textwidth]{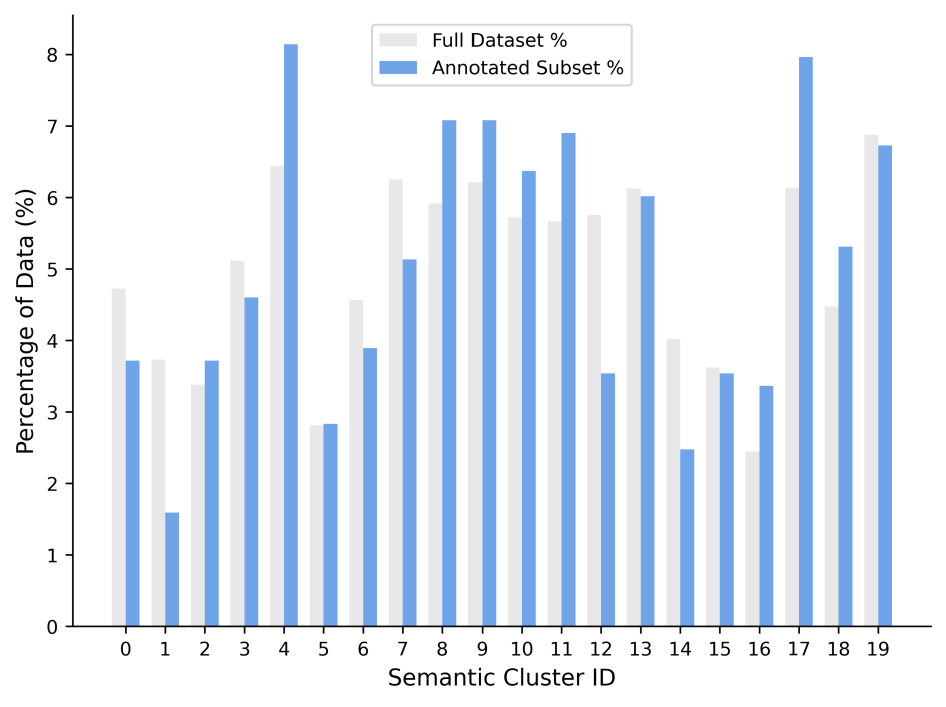}
       \caption{Cluster distribution comparison}
   \end{subfigure}
   \caption{Semantic Representativeness Analysis with k=20 Clusters (Elbow Method).}
   \label{fig:sample-analysis}
\end{figure*}
Given the concise nature of the claims in our dataset and their associated evidence, the validation task required a carefully designed approach. Similar to the previous tasks, the validation LLM was guided by a careful COT prompt (see the supplementary material for the full prompt). Our prompt was inspired by how humans perform evaluation tasks, which involve inductive reasoning and a focus on data‐driven analysis that is consistent with cognitive models of human information processing \cite{kahneman2011thinking}. Our COT prompt thus mimicked, to some extent, a simplified reasoning framework of a human evaluator, thereby encouraging the LLM to process information in a more methodical and verifiable manner, leading to higher accuracy and better transparency of the output. To streamline this process and make it more manageable for the validation LLM, we divided the validation task into two sub-tasks, one for supported claims and one for refuted ones, rather than addressing all three cases at once (i.e. supported, refuted or NEI). This is similar in spirit to breaking down a three-way classification problem into two discrete binary classification ones. Thus, the validation LLM was provided with the set of supported claims and asked to decide whether each claim is indeed supported by its evidence, or whether it needs to adjust its label to NEI. The same procedure was carried out on the refuted claims generated by the refutation task. 

Building on this framework of human-inspired reasoning, the task of claim validation was broken down into simple systematic steps assessing different facets of the claim-evidence relationship. The first step examined entity mapping between the claim and the evidence, and their coreference alignment. In instances where the mapping was deemed ambiguous or revealed significant discrepancies, the label was adjusted to NEI, \emph{and the subsequent steps were skipped}. In the second step, the validation LLM assessed the quality of the linguistic style of both the claim and the evidence, including, but not limited to, verb-tense alignment, gender and number agreement patterns, and the consistent use of voice across statements. We also emphasized the importance of the complex system of Arabic pronouns and their coreferences. The LLM then produced a quantitative linguistic score ranging between 0 and 100. In the third step, it evaluated the vocabulary alignment by inspecting important terms and phrases such as the qualifiers and provided an alignment score ranging between 0 and 100. The fourth and final step focused on semantic similarity. The LLM determined the semantic coherence, which required taking scope, temporality and modality into account while meticulously examining gaps in focus, subject–predicate relations, or certainty levels and produced a similarity score that ranges between 0 and 100. The linguistic, vocabulary, and semantic scores were aggregated into one comprehensive score that weighted the individual scores by 30\%, 30\%, and 40\%, respectively. Any claim-evidence pair with an aggregated score less than 88 was labeled as NEI. 

\begin{figure*}[t!]
   \centering
   \begin{subfigure}[b]{0.46\textwidth}
       \centering
       \includegraphics[width=\textwidth]{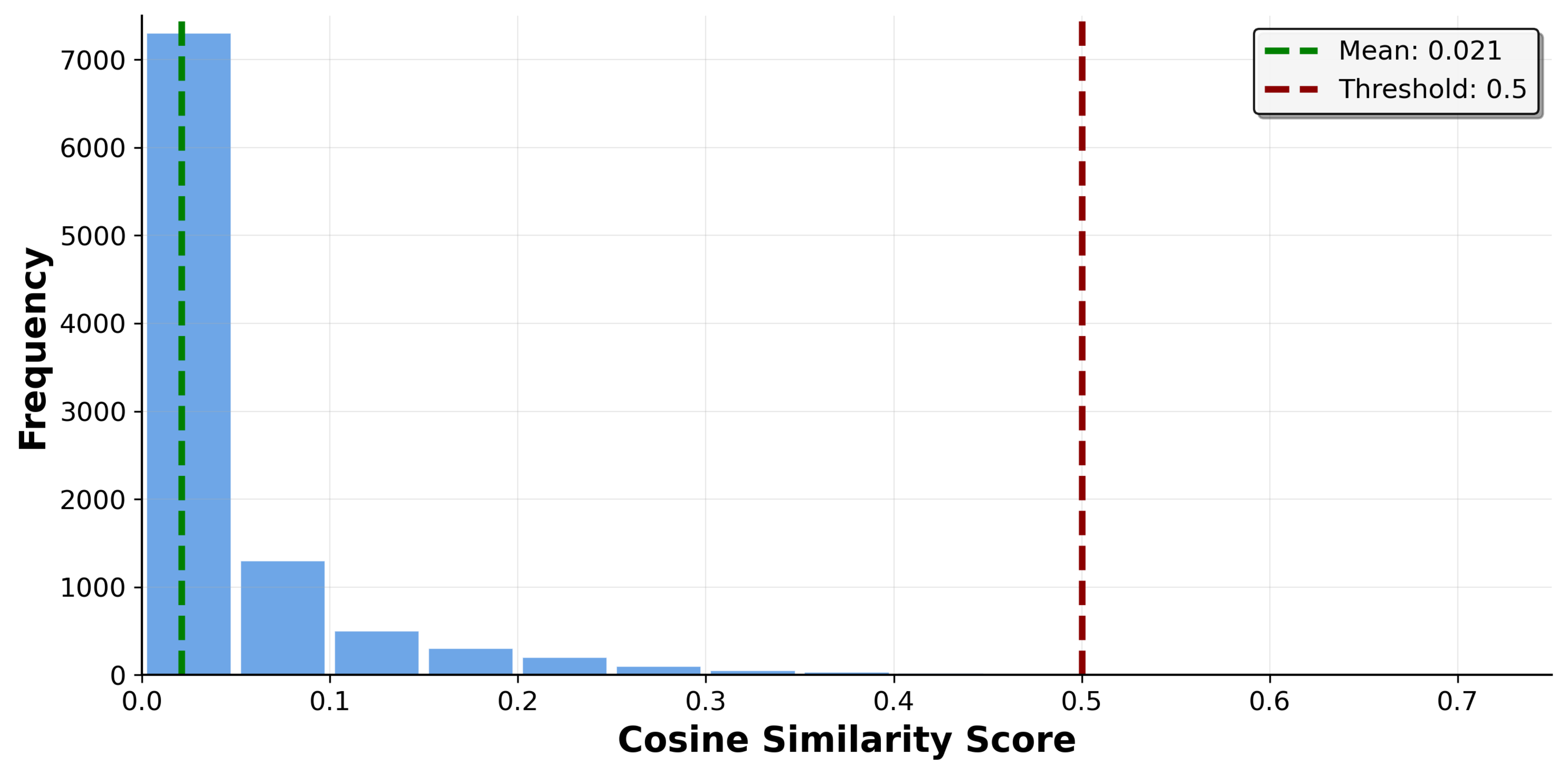}
       \caption{Claim-to-Claim Cosine Similarity}
       \label{fig:claim-claim-sim}
   \end{subfigure}
   \hfill
   \begin{subfigure}[b]{0.46\textwidth}
       \centering
       \includegraphics[width=\textwidth]{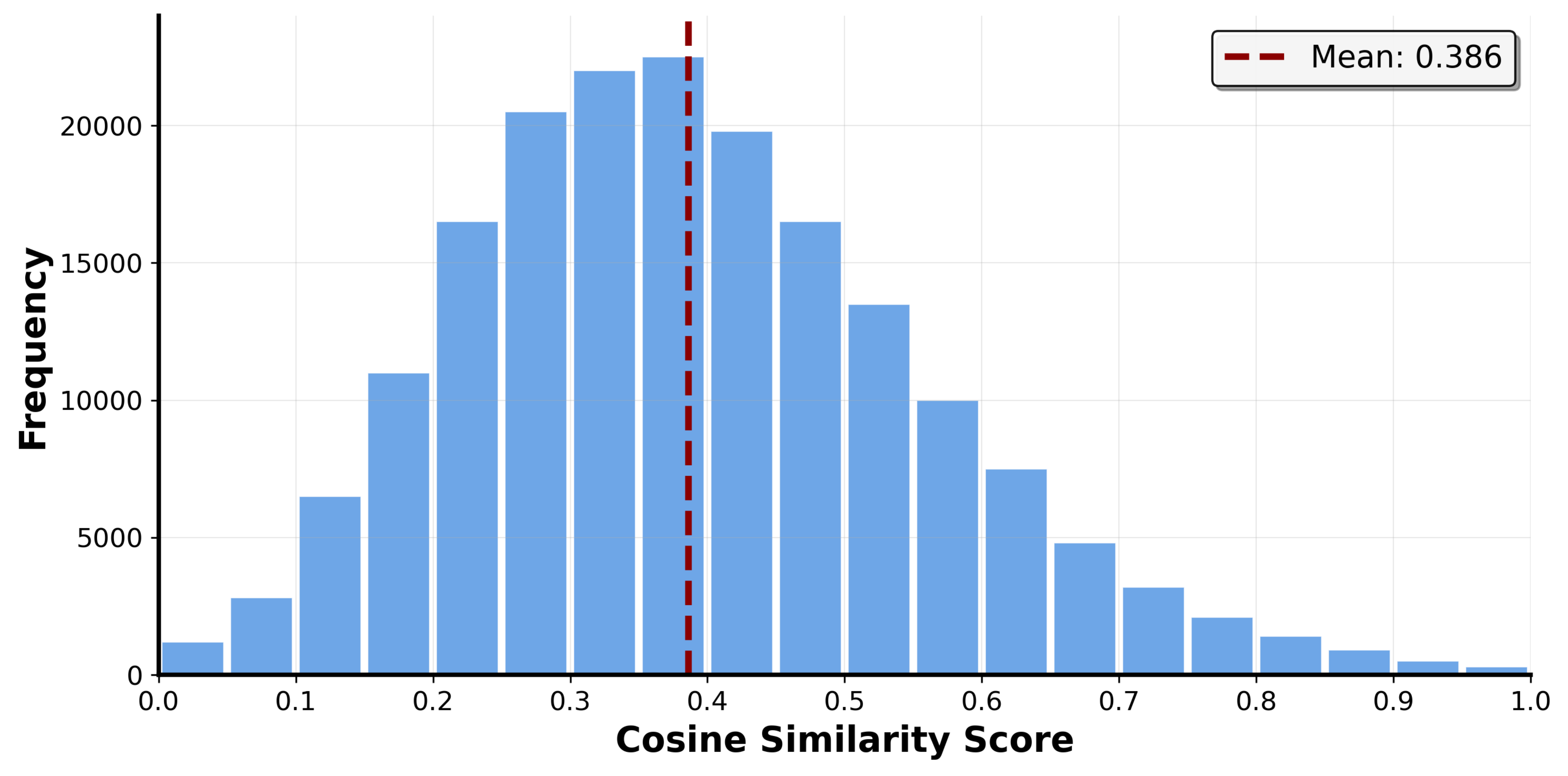}
       \caption{Claim-to-Evidence Cosine Similarity}
       \label{fig:claim-evidence-sim}
   \end{subfigure}
   \caption{Distribution of Cosine Similarity Scores in \dataset}
   \label{fig:similarity-distributions}
\end{figure*}
We experimented with various LLMs for the validation task. To be able to assess their accuracy, we relied on human annotations to generate ground-truth claim-evidence pairs. To this end, we randomly sampled 400 claim-evidence pairs from the dataset (200 supported, 200 refuted). We then conducted a semantic coverage analysis to ensure that this subset captured the topical diversity of the full dataset (Figure \ref{fig:sample-analysis}). Specifically, we projected both the annotated sample and the entire dataset of 182k pairs into a latent semantic space using multilingual-e5-large embeddings \cite{wang2022text} combined with Uniform Manifold Approximation and Projection (UMAP) \cite{mcinnes2018umap}, and applied K-Means clustering with $k=20$ determined via the elbow method. As illustrated in Figure \ref{fig:sample-analysis}, the 400 human-annotated samples achieve 100\% coverage across all 20 semantic clusters. To assess whether the sample is representative of the full dataset, we applied two distributional similarity tests. The Jensen–Shannon Divergence indicates a close match between the two distributions ($\mathrm{JSD} = 0.0879$). In addition, Kolmogorov–Smirnov tests on spatial coordinates show no significant differences (X: $\mathrm{KS} = 0.047$, $\mathrm{p} = 0.157$; Y: $\mathrm{KS} = 0.042$, $\mathrm{p} = 0.266$). This indicates the samples are not clustered in specific areas but occupy the semantic space uniformly. All in all, these measures confirm that the expert-annotated subset covers the distributional features and diversity of the global dataset.

With the representativeness of the sample confirmed, we then carried out expert annotation. Two native Arabic-speaking \emph{expert} annotators, one of whom is the lead author of this paper and another who is a PhD student, \emph{collectively} examined each claim-evidence pair closely. With these 400 ground-truth claim-evidence pairs, we evaluated five different LLMs in terms of their validation accuracy. Table \ref{tab:eval} summarizes their performance: Claude Sonnet 3.5 led with an accuracy of 86\% with respect to ground-truth \emph{supported} claim-evidence pairs, followed by DeepSeek R1 (80\%), GPT-4o (82\%), and Llama 3.1 70b at approximately 71\%. For \emph{refuted} claims, Claude Sonnet 3.5 led with an accuracy of 88\%, as can be seen in Table \ref{tab:eval}. Based on these results, we selected Claude Sonnet 3.5 as the final validation LLM, which was then used to validate all the claim-evidence pairs in \dataset, and to adjust their labels to NEI if needed.

To be able to compare the performance of the validation LLM to human-level performance, the ground-truth claim-evidence pairs were also annotated by four human annotators, \emph{different} from the expert annotators. All annotators were university graduates and native Arabic speakers. The annotators received training from expert annotators to follow the same structured guidelines that were used to design the LLM validation framework. These guidelines required a stepwise assessment of entity and coreference alignment, linguistic alignment, vocabulary and qualifier alignment, and overall semantic consistency, with ambiguous cases labeled as NEI and no external knowledge allowed. The annotators were then divided into two teams (two members per team), one responsible for validating the ground-truth supported claims, and one for refuted claims. For the supported pairs, the inter-annotator agreement between the two non-expert annotators as measured by Cohen’s kappa was $\kappa = 0.89$. The first non-expert annotator achieved an agreement of $\kappa =0.86$ with the expert annotators, and the second annotator obtained an agreement of $\kappa =0.80$. Similarly, for the refuted pairs, the inter-annotator agreement between the two non-expert annotators was $\kappa = 0.94$, with the first non-expert annotator achieving an agreement of $\kappa =0.82$ with the expert annotators, and the second achieving $\kappa =0.84$. This highlights that using an LLM for validation yields comparable accuracy to human validation. On the other hand, human annotators were only able to validate an average of 60 claim-evidence pairs per hour, which clearly demonstrates the necessity of using an automated validation technique to be able to validate the generated claim-evidence pairs in \dataset, which consists of over 180,000 claim-evidence pairs. 

\begin{table*}[t]
\centering
\small
\caption{Validation Task Accuracy for Supported and Refuted Claims}
\label{tab:eval}
\begin{tabularx}{\textwidth}{l*{5}{>{\centering\arraybackslash}X}}
\toprule
\textbf{Claim Type} & \textbf{Llama 3.1 70b} & \textbf{Llama 3.1 8b} & \textbf{GPT-4o} & \textbf{Claude Sonnet 3.5} & \textbf{DeepSeek R1} \\
\midrule
Supported/NEI & 71\% & 38\% & 82\% & \textbf{86\%} & 80\% \\
Refuted/NEI   & 78\% & 43\% & 85\% & \textbf{88\%} & 84\% \\
\bottomrule
\end{tabularx}
\end{table*}
\section{Dataset Analysis}
\dataset was generated using the rigorous framework described in the previous section. It consists of \emph{181,976} claims that are accompanied by textual evidence that either supports or refutes the claims, or is insufficient to judge whether the claims are true or false (i.e. not enough information, or NEI). The claims in \dataset are unevenly distributed across the three classes, with the majority of the claims belonging to the \textit{supported} class ($111,303$ instances), followed by \textit{refuted} ($42,109$) and \textit{NEI} ($28,564$). This can be attributed to various factors. First, the claim generation task is relatively easier since it involves extracting evidence from a Wikipedia article chunk, followed by generating a claim that is supported by the extracted evidence. The claim refutation task, on the other hand, is more complex, as it involves generating false claims from extracted evidence, a task that has proven to be more difficult even for humans \cite{thorne2018fever}. Finally, NEI claims were generated through the validation task, which corrected any mislabeled supported or refuted claims. Given the rigorous approaches used for both claim generation and claim refutation, only a small subset of the claim labels were adjusted to NEI through the validation task.

\begin{table*}[t!]
\centering
\small
\caption{Performance of Fine-Tuned Models Using \dataset}
\label{tab:treval}
\begin{tabularx}{\textwidth}{l*{5}{>{\centering\arraybackslash}X}}
\toprule
\textbf{Model} & \textbf{Accuracy} & \textbf{F1-Macro} & \textbf{F1-Supported} & \textbf{F1-Refuted} & \textbf{F1-NEI} \\
\midrule
AraBERTv2              & \textbf{82\%} & 75\%          & \corr{88\%}          & 86\%          & 51\%          \\
AraModernBert-Base-V1.0 & 81\%          & \textbf{77\%} & 87\%                 & 91\%          & \corr{\textbf{52\%}} \\
Llama-3.1-8B           & 79\%          & 75\%          & \corr{85\%}          & \textbf{93\%} & \corr{47\%}          \\
Qwen2.5-7B             & \textbf{82\%} & 76\%          & \corr{\textbf{89\%}} & 88\%          & \corr{51\%}          \\
\bottomrule
\end{tabularx}
\end{table*}
To demonstrate the diversity of \dataset, we computed the similarity between its claims and claim–evidence pairs. Figure \ref{fig:claim-claim-sim} presents the similarity scores between claims, revealing a highly skewed distribution with a mean cosine similarity of 0.021, which indicates minimal lexical overlap between claims and highlights the dataset’s coverage of diverse factual assertions rather than highly similar claims. Similarly, Figure \ref{fig:claim-evidence-sim} shows the similarity scores between a claim and its corresponding evidence, yielding a balanced distribution centered around 0.386, which indicates that while claims and their evidence are lexically coherent, they are nonetheless meaningfully distinct. This confirms that the claim–evidence pairs are sufficiently complex to support the development of robust fact-checking approaches that move beyond basic text matching and instead require more advanced logical reasoning. Further analysis of the claim-evidence pairs reveals that the claims are consistently concise with an average length of 15.41 words (median: 15.00). Similarly, the evidence  associated with the claims is equally concise, with the majority of the evidence (89.4\%) consisting of only a single sentence, and only 10.6\% consisting of multiple sentences. 

Finally, to demonstrate the utility of \dataset in developing robust automatic Arabic fact-checking models, we fine-tuned four state-of-the-art transformer-based models to perform automatic fact-checking using \dataset. All experiments were performed on a high-performance computing cluster equipped with three NVIDIA RTX A6000 GPUs. The first two models, AraBERTv2 \cite{antoun2020arabert} and AraModernBert-Base-V1.0 \cite{elshehy2026aramodernbert}, are encoder-based models and were fine-tuned to perform sequence classification by concatenating claims and their evidence with “[SEP]” separator tokens as input, and the labels (\textit{supported}, \textit{refuted}, and \textit{NEI}) as target classes. The two models were tuned using maximum sequence lengths of 512 and 1024 tokens, respectively, a batch size of 32 per GPU, a learning rate of 2e-5, and were trained for 3 epochs using the AdamW optimizer \cite{loshchilov2017decoupled} with linear warmup.  In addition to the encoder-based models, we also fine-tuned two generative models, Llama-3.1-8B \cite{grattafiori2024llama} and Qwen2.5-7B \cite{team2024qwen2}. The generative models were fine-tuned via Parameter-Efficient Fine-Tuning (PEFT) using Low-Rank Adaptation (LoRA) \cite{hu2022lora} with LoRA rank r=64, alpha=128, and dropout=0.1. The two models had a maximum sequence length of 2048 tokens, effective batch size of 33 across \gc{three} GPUs through gradient accumulation \gc{(a per-GPU batch size of 1 with 11 accumulation steps)}, learning rate of 2e-4 \gc{with a cosine schedule, 3\% warmup and a weight decay of 0.01}, and were trained for 3 epochs using instruction-following prompts in the Llama chat format. \gc{The 400 human-annotated claim-evidence pairs used as the test set were first removed from \dataset, so that none of them were seen during training or model selection. All four models were then trained using the remaining pairs}, which were split into 90\% for training and 10\% for validation. The models were tested using the human-annotated test set (consisting of the 400 ground-truth claim-evidence pairs) that was used for the claim validation task described in the previous section. \gc{For the encoder models, m}odel selection was based on macro F1-score using the validation set\gc{, while for the generative models the checkpoint obtained at the end of the third epoch was used}, and final evaluation metrics were computed using standard classification measures including per-class F1-scores and overall accuracy.

Table \ref{tab:treval} shows the performance of the four fine-tuned models using \dataset. As can be seen from the table, AraBERTv2 and Qwen2.5-7B achieved the highest overall accuracy (82\%), while AraModernBert-Base-V1.0 exhibited the most balanced performance, attaining the highest macro F1-score (77\%). Breaking down the performance by class, all four models were quite successful in detecting refuted claims with the worst model, AraBERTv2, achieving an F1-score of 86\% for refuted claims. On the other hand, \corr{the performance across the other two classes (supported and NEI) is consistent between encoder-based and generative models. All four models} excel in verifying supported claims, achieving F1-scores of \corr{85–89\%, with Qwen2.5-7B achieving the highest F1-score (89\%)}. In contrast, \corr{all four models find it considerably more difficult to identify} cases where the evidence lacks sufficient information to support the claims (i.e., NEI class), \corr{achieving F1-scores of only 47–52\%} in this category. Notably, Llama-3.1-8B achieves exceptional performance on \textit{refuted} claims (93\% F1-score) \corr{but obtains the lowest F1-score on the NEI class (47\%)}. Due to computational constraints and limited budget for training open-source large language models, we employed parameter-efficient fine-tuning using LoRA rather than full model fine-tuning. This approach, while computationally feasible, may preserve pre-trained biases and limit \corr{the performance of the generative models}. \corr{Since all four models share the same weakness on the NEI class, improving the detection of claims whose evidence does not provide enough information to judge their validity remains the main challenge for automatic fact-checking using \dataset.}

\section{\texorpdfstring{Limitations}{Limitations}}
\label{chap:limitation}
\pagestyle{plain}
Our work represents a significant advancement by addressing the scarcity of resources for Arabic fact-checking. \dataset is derived exclusively from Arabic Wikipedia to align with FEVER, ensuring a comparable benchmark for Modern Standard Arabic (MSA). Even though it is a centralized and reliable knowledge base, Wikipedia is still vulnerable to crowd-sourced biases and/or inaccuracies. Consequently, \dataset may not fully reflect the landscape of real-time misinformation found in news and social media. However, it is important to note that the described framework is developed to be source-agnostic and can be generalized and used to generate new domain-specific datasets in the future.
In the linguistic scope, we consciously and by design for this paper targeted Modern Standard Arabic (MSA), as it is the official lingua-franca of the Arab world, used in education, formal media and legislation. As a result, \dataset may not capture the full complexities of Dialectal Arabic, but can later be extended to it by adjusting the linguistic constraints within the prompt instructions.
Finally, \dataset is a synthetic dataset generated by LLMs; this method allows for scale and time efficiency that is impossible with manual annotation. However, this introduces the risk of propagating model-specific artifacts or hallucinations. We mitigated this as much as possible through a rigorous validation task, which achieved comparable performance to human annotators. Nevertheless, the complexity of the generated claims is tightly bounded by the capabilities of the employed LLM especially when processing morphologically complex languages like Arabic.

\section{Conclusion}
\label{chap:conclusion}
\pagestyle{plain}

We presented \dataset, a comprehensive and large Arabic fact-checking dataset, fully generated using large language models (LLMs) through a rigorous pipeline consisting of three interleaved tasks, namely claim generation, claim refutation and claim validation. The dataset consists of 181,976 claims, each of which is associated with textual evidence that either supports or refutes it. In cases where the evidence is not sufficient to judge the validity of a claim, the pair is then labeled as not enough information. The dataset was thoroughly validated using an LLM, achieving an accuracy of 86\% for supported claims, and 88\% for refuted claims on ground-truth human-annotated claim-evidence pairs.
In addition to \dataset serving as the largest general-purpose Arabic fact-checking dataset that can be used for training automatic fact-checking models or for evaluating them, our dataset generation framework serves as a general framework that can be used to generate similar datasets for other low-resource languages. The dataset and the code to implement the framework, including the different prompts used to instruct the LLMs to perform each task, are publicly available to advance automatic fact-checking efforts in Arabic and other languages. 

In future work, we plan to augment \dataset with claims generated from additional sources other than Wikipedia and include various Arabic dialects, extending beyond Modern Standard Arabic. Finally, we plan to fine-tune small and large language models for Arabic fact-checking, and compare them to different LLM-based fact-checking approaches.


\backmatter

\bmhead{Acknowledgments}

We would like to thank the expert annotator Aya Mourad, as well as the other four human annotators for their valuable contribution to the human evaluation of the dataset. 

\section*{Declarations}

\begin{itemize}
\item Funding: This work was funded by the American University of Beirut Research Board (URB), Award number 104518.
\item Conflict of interest/Competing interests: The authors have no conflicts of interest to declare that are relevant to the content of this article.
\item Ethics approval and consent to participate: The authors declare that this work does not require any ethics approval. All human evaluations were voluntarily done.
\item Consent for publication: All authors have provided consent for publication.
\item Data availability: All data is publicly available at \url{https://zenodo.org/records/15020544}.
\item Materials availability: All materials are publicly available at \url{https://github.com/chriskhalil/ARAFA}.
\item Code availability: All code is publicly available at \url{https://github.com/chriskhalil/ARAFA}.
\item Author contribution: Khalil (Conceptualization, Methodology, Implementation, Experiments, Writing), Elbassuoni (Conceptualization, Supervision, Methodology, Writing), Assaf (Conceptualization, Methodology, Supervision). 
\end{itemize}

\noindent






\bibliography{sn-bibliography}




\end{document}